\documentclass{article}

\usepackage[preprint]{corl_2026} 
\usepackage{graphicx} 
\usepackage{capt-of}
\usepackage{multirow} 
\usepackage{times}
\usepackage{etoolbox}
\usepackage{latexsym}
\usepackage{array}
\usepackage[T1]{fontenc}
\usepackage[utf8]{inputenc}
\usepackage{microtype}
\usepackage{inconsolata}
\usepackage{booktabs}
\usepackage{hyperref}
\usepackage{amsmath}
\usepackage{enumitem}

\newcommand{\name}{\textsc{VLAlert}}

\title{Observe Before You Alert: Adaptive Driver Alerting with Vision–Language Models}

\author{
  Yuhang Wang\\
  Department of Civil Engineering\\
  University of South Florida 
  United States\\
  \texttt{yuhangw@usf.edu} \\
  \And
  Lingyao Li\\
  School of Information\\
  University of South Florida 
  United States\\
  \texttt{lingyaol@usf.edu} \\
  \And
  Hao Zhou\\
  Department of Civil Engineering\\
  University of South Florida 
  United States\\
  \texttt{haozhou1@usf.edu} \\
}

\makeatletter
\let\@makespecialcolbox\@make@specialcolbox
\makeatother

\begin{document}
\maketitle




\begin{abstract}
Driver alerting from dashcam video requires sequential decision-making under partial observability: a system must decide not only whether a scene is risky, but also when the evidence is sufficient to warn. Most existing accident anticipation models output a binary risk score, leaving ambiguous scenes to be handled by thresholding. We propose \name{}, a vision--language alerting framework that casts warning generation as a tri-action policy over \textsc{Silent}, \textsc{Observe}, and \textsc{Alert}. The \textsc{Observe} action acts as an internal evidence-gathering decision that delays uncertain warnings and changes the next observation window, creating a lightweight perception--action loop for adaptive alerting. \name{} uses Qwen3-VL-4B as a safety-evidence generator and pools hidden states from structured belief spans to form compact representations for danger estimation and policy prediction. We evaluate \name{} on \name{}-Bench, a unified per-tick benchmark from four real-world dashcam alert datasets, and further test transfer to held-out naturalistic ADAS takeover clips. On \name{}-Bench validation, \name{} achieves the highest deployment-oriented utility among tested baselines, with DAUS 0.4878 compared with 0.4752 for Open-BADAS, and improves AUROC, AP\textsubscript{tick}, F1\textsubscript{t}, and balanced accuracy from 0.610, 0.176, 0.276, and 0.581 to 0.689, 0.195, 0.297, and 0.648, respectively. On 221 held-out ADAS-TO-Critic clips, \name{} improves R@5s from 74.2\% to 88.7\% and F1 from 0.585 to 0.686. These results indicate that adaptive observation and safety-focused VLM representations provide measurable gains for driver-facing alert decisions.

\end{abstract}


\begin{figure}[!ht]
    \centering
    \includegraphics[width=1.0\linewidth]{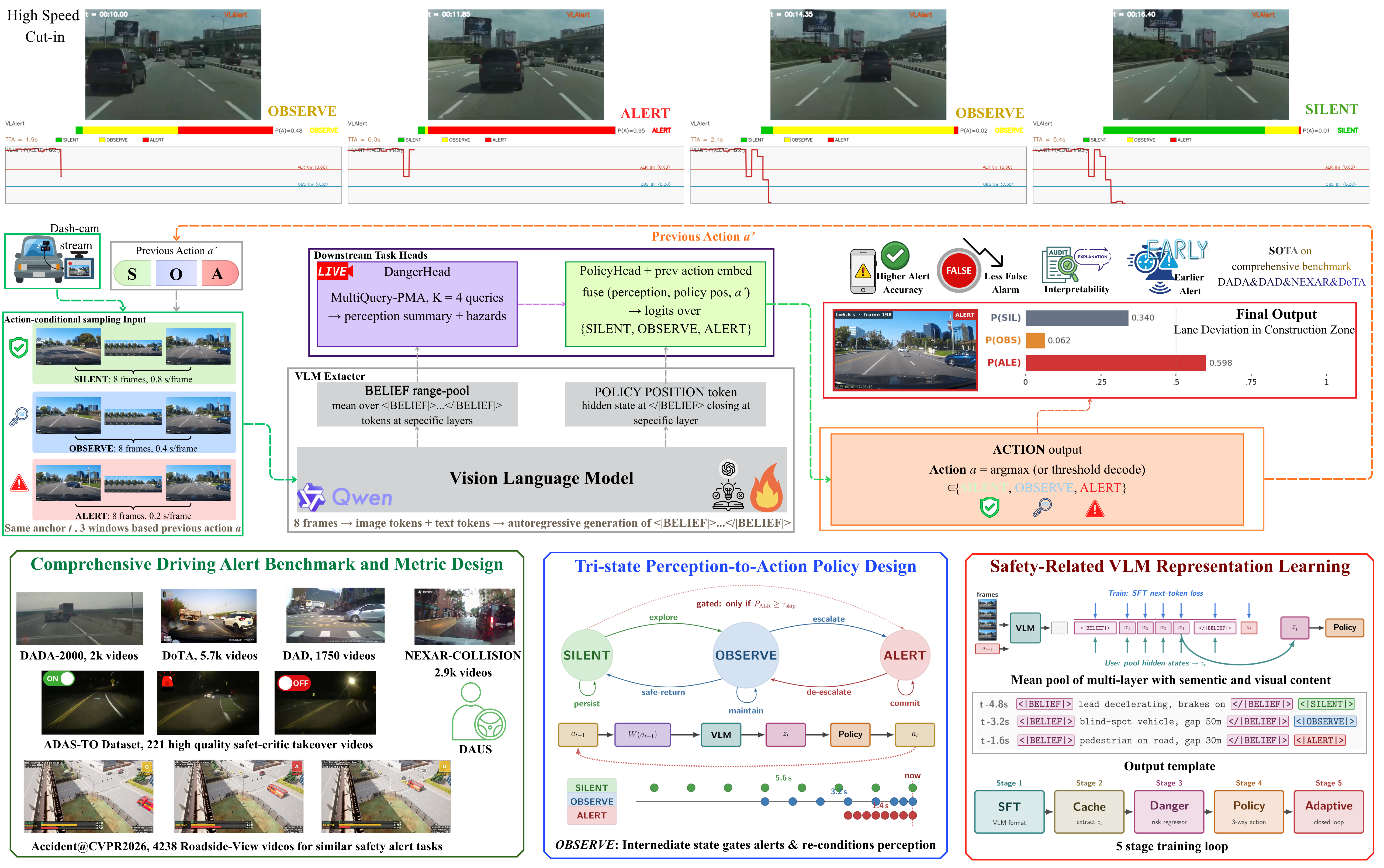}
    \caption{Overview of \name. }
    \vspace{-8mm}
    \label{fig:teaser}
\end{figure}

\section{Introduction}

Road traffic crashes remain a major safety concern. Many safety-critical events are associated with delayed hazard perception, driver distraction, or limited time for corrective action once a conflict becomes imminent~\cite{world2024global,NCSA2026DistractedDriving,qu2025examining,omran2023driving}. Dashcam-based driving alert systems aim to reduce this response gap: a camera mounted on the windshield or dashboard observes the vehicle's first-person road scene and warns the driver before a hazardous situation becomes difficult to avoid~\cite{moura2025nexar,zhao2023effects,yu2022personalized}. Such warnings can support braking and steering in human driving, and takeover decisions when assisted driving is active~\cite{huang2022takeover,tan2022effects}. However, their value depends not only on detecting visual risk, but also on whether the warning is issued at the right time: early enough to support action, reliable enough to be trusted \cite{WANG2025108022}, and restrained enough to avoid unnecessary interruptions \cite{ZHANG2025108008}. This suggests that driver alerting should be treated not only as visual risk recognition, but also as a temporally grounded decision problem that requires interpreting evolving traffic context and determining when the evidence is sufficient to warn~\cite{10.1145/3664647.3680672,liao2024and,zhang2025eyes, liu2026predict}.

Recent work in driver alert system has gone beyond generic visual classification toward learned risk estimation from richer video and multimodal signals. Multimodal accident-understanding datasets and VLM-based analysis frameworks now use language to describe accident causes, relevant agents, prevention cues, and visual grounding~\cite{10657507, zhang2025language}. At the same time, video foundation and world-model representations have been adapted to ego-centric collision prediction, while self-supervised risk objectives can impose temporal structure on the danger signal, such as the assumption that danger should generally increase as an accident approaches~\cite{assran2025v,goldshmidt2025badas,pjetri2024self}. Complementary work on naturalistic driving further suggests that useful collision-risk signals can be learned from large-scale ordinary interactions without dense manual crash-risk labels~\cite{jiao2026learning}. Together, these developments make it increasingly plausible to train alert systems from visual, textual, and naturalistic evidence rather than relying only on manually designed surrogate thresholds.

These advances also leave several open questions for driver alerting. First, existing evaluations are still largely built around accident-centric datasets and ranking metrics, which are useful for anticipation but provide limited evidence about nuisance alerts, normal-driving behavior, or alignment with human-perceived risk~\cite{zhao2026accident,zhao2024driver}. Second, most models reduce risk to a binary prediction or scalar score; thresholds can control sensitivity, but they do not provide an explicit mechanism for deferring a warning and gathering more evidence when the scene is ambiguous~\cite{11295937,xie2025vlms}. These gaps motivate a unified alerting benchmark, an observation-aware alert policy, and a VLM interface that extracts alert-relevant representations.

To address these gaps, we propose \name, a vision--language framework for adaptive driver alerting system. \name{} adapts a VLM as a safety-evidence generator and uses the hidden states inside the generated evidence spans as compact features for danger estimation and policy prediction. Instead of reducing alerting to a binary decision or a thresholded risk score, \name{} uses a tri-action policy over {Silent, Observe, Alert}, where Observe allows the system to defer, and gather more evidence before issuing an uncertain warning.

Our contributions are threefold: i) we build  \name-Bench, a unified per-tick benchmark that introduces DAUS to jointly measure alert coverage, false-alert burden, and lead time and combines multiple driving-alert datasets that features a new real-world ADAS takeover collection to test whether the learned model aligns with human risk perception and reaction; ii) we formulate alerting as an observation-aware policy over {Silent, Observe, Alert}, where Observe is supervised by risk-cue onset annotations and defers uncertain alerts before committing; iii) we train a VLM to generate structured safety-evidence spans and pool their hidden states as alert-relevant features, reducing dilution from generic scene description and improving downstream alert prediction.

\section{Related Work}

Early accident anticipation research is shaped by dashcam datasets such as DAD~\citep{chan2016anticipating} and DADA-2000~\citep{fang2019dada,fang2021dada}, which focus on predicting whether and when a crash occurs. DoTA~\citep{yao2022dota} extends this setting to traffic anomaly detection with richer anomaly categories. More recent benchmarks move beyond crash occurrence alone. For example, MM-AU \cite{10657507} provides temporally aligned language annotations, object boxes, accident reasons, and prevention descriptions for causal and semantic accident understanding~\citep{10657507}. NEXAR-Collision focuses on ego-centric collision and near-collision prediction, with alert-time annotations and AP-based evaluation across pre-event intervals~\citep{moura2025nexar}.

A major line of work improves accident anticipation through specialized perception, attention, and fusion designs. Recent works have used depth-enhanced 3D scene modeling~\citep{liao2024real}, context-aware and temporal-focus attention~\citep{liao2024crash}, LLM-assisted accident localization across when, where, and what dimensions~\citep{liao2024and}, and driver-attention-guided transformers for identifying risky traffic participants~\citep{kumamoto2025aat}. A complementary direction is data-centric end-to-end collision prediction. BADAS and Open-BADAS/BADAS-2.0 use V-JEPA2-style video foundation models and large ego-centric dashcam corpora to predict threats involving the recording vehicle~\citep{goldshmidt2025badas,goldshmidt2026beyond,assran2025v}.

\paragraph{Surrogate safety measures and naturalistic risk learning.}
Classical driver warning systems often rely on surrogate safety measures such as TTC, THW, post-encroachment time, or spacing-based conflict indicators. These measures are interpretable and useful in structured settings, especially car-following and lane-change scenarios, but they require hand-designed thresholds and may not generalize across interaction types, scene layouts, weather, lighting, or occlusion. Recent work learns more flexible conflict measures from trajectory data. For example, Jiao et al. model traffic conflicts as context-dependent extreme events~\citep{jiao2025unified} and study missed and false alarms in vehicle-spacing-based conflict detection~\citep{jiao2024minimising}. GSSM further shows that large-scale naturalistic driving data can reveal unsafe deviations even without crash labels~\citep{jiao2026learning}.



\section{Method}
\label{sec:method}
We present \name, a VLM framework for streaming driver alerting from ego-view video.
At each decision tick,~\name~receives an 8-frame observation window and predicts a policy over SILENT, OBSERVE, and ALERT.
The method consists of four parts: a belief-state formulation, a unified alert benchmark, a VLM-based alert architecture, and a closed-loop training procedure.


\subsection{Problem Formulation}
\label{sec:method_formulation}
Driver alerting is a partially observed sequential decision problem.
At tick $t$, the system observes only an ego-view video history, while safety-critical factors such as agent intent, occluded objects, road affordances, and driver response remain latent.
We therefore use a POMDP-inspired formulation~\citep{lauri2022partially}, but do not explicitly model transition or observation probabilities.
Instead,~\name~learns an amortized belief representation from video and trains a policy on top of it.
The action space is
\begin{equation}
\label{eq:action_space}
\mathcal{A}=\{SILENT,OBSERVE,ALERT\}.
\end{equation}
SILENT suppresses warning.
OBSERVE defers commitment and requests more evidence.
ALERT triggers a driver-facing warning.
Unlike binary collision predictors,~\name~treats warning as a sequential policy decision under uncertainty.
Let $V_{\leq t}$ denote the video stream up to tick $t$ and let $a_{t-1}$ denote the previous action.
A fixed temporal window forces all risk states to use the same sampling scale.
This is undesirable because long windows provide driving context and reduce false alarms, while dense short windows preserve fast motion cues needed for imminent warnings.
Recent collision-prediction results show that window length changes precision, recall, and alert timing, since overly short windows miss developing hazards and overly long windows include non-indicative frames~\citep{goldshmidt2025badas}.
\name resolves this trade-off with an action-conditioned sampler:
\begin{equation}
\label{eq:adaptive_window}
X_t=W(V_{\leq t},a_{t-1})=\{I_{t,1},I_{t,2},\ldots,I_{t,8}\}.
\end{equation}
The sampler uses a wide sparse window after SILENT, a medium dual-resolution window after OBSERVE, and a dense short-range window after ALERT.
Thus, the action does not merely emit a decision; it also re-targets the next visual observation.
Given $X_t$,~\name~computes a learned belief embedding and predicts the next action:
\begin{equation}
\label{eq:belief_policy}
\hat b_t=\Phi_\theta(X_t), \qquad \pi_t=\pi_\psi(a_t\mid \hat b_t,a_{t-1}).
\end{equation}
Here $\hat b_t$ is not an explicit posterior over latent driving states.
It is a learned hidden-state representation that summarizes task-relevant evidence under partial observability.
This formulation gives OBSERVE an operational role as an evidence-gathering action rather than an intermediate confidence bin.

\begin{figure}[t]
  \centering
  \includegraphics[width=1\linewidth]{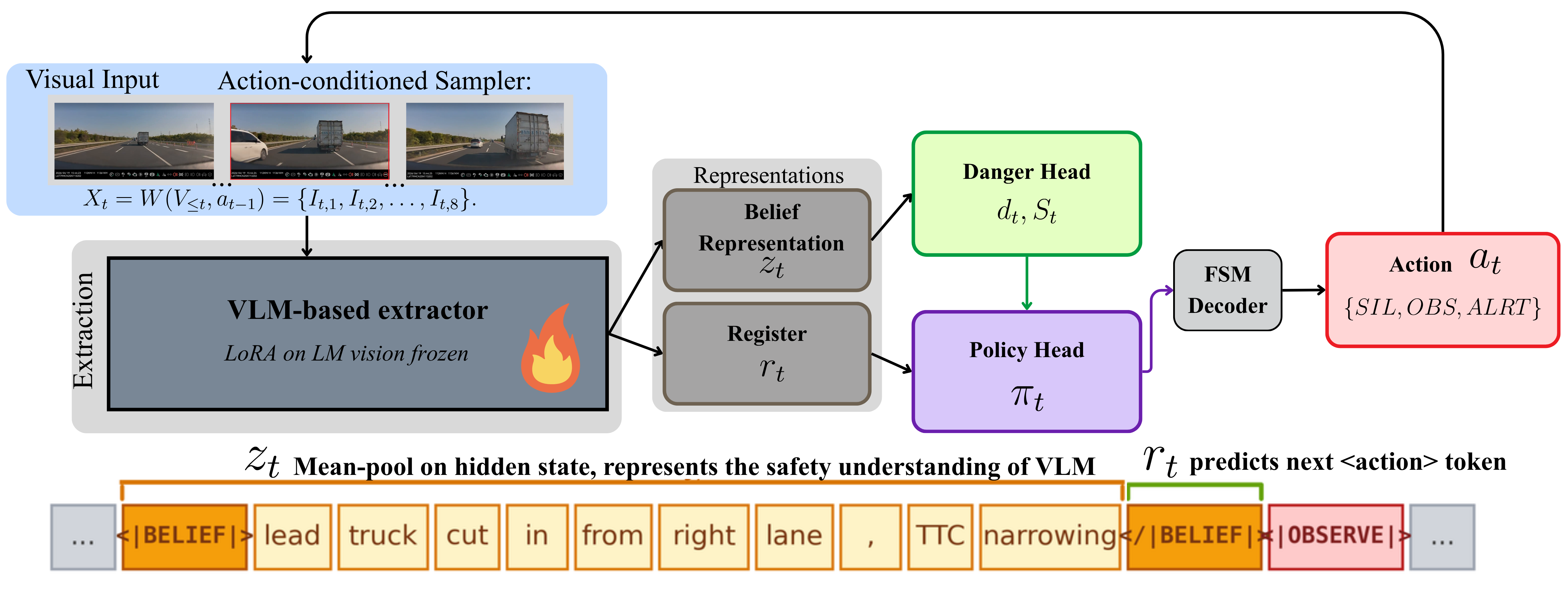}
  \caption{\textbf{Architecture of \name.}
 ~\name~ extracts belief representations from ego-view video with a VLM and predicts a tri-policy over $\{SILENT, OBSERVE, ALERT\}$.
  }
  \label{fig:framework}
\end{figure}

\subsection{Benchmark Construction}
\label{sec:method_benchmark}

We construct \textbf{\name-Bench}, a unified per-tick benchmark for driver-facing alert decisions.
The benchmark pools four real-world dashcam corpora, Nexar Collision~\citep{moura2025nexar}, DoTA, DAD, and DADA-2000, for in-domain training and validation.
Two additional sources are held out for test-only evaluation: ADAS-TO-Critic~\citep{wang2026adas}, which contains expert-reviewed human takeover events from production ADAS logs, and ACCIDENT~\citep{accidentcvpr2026}, a synthetic collision set for OOD testing.

All sources are converted to a common 1\,Hz streaming format.
Each tick contains an 8-frame observation window and one action label in
$\{\textsc{Silent},\textsc{Observe},\textsc{Alert}\}$.
Let $t_f$ denote the timestamp of the tick anchor frame, $t^\star$ the event time
(collision, anomaly peak, or takeover, depending on the source), and
$t^\circ\!\le\! t^\star$ the first timestamp at which a source-specific
risk-cue annotation marks the developing hazard.
For collision-style corpora, the tick action label is
\begin{equation}
\label{eq:label_rule}
y_f =
\begin{cases}
\textsc{Alert},   & t^\star - L_{\textsc{a}} \;\leq\; t_f \;<\; t^\star, \\[2pt]
\textsc{Observe}, & t^\circ \;\leq\; t_f \;<\; t^\star - L_{\textsc{a}}, \\[2pt]
\textsc{Silent},  & t_f \;<\; t^\circ, \\[2pt]
\textit{discard}, & t_f \;\geq\; t^\star,
\end{cases}
\end{equation}
with $L_{\textsc{a}}=5$\,s unless otherwise specified.
The \textsc{Alert} label therefore corresponds to the pre-event interval in which a driver-facing warning is expected to be actionable.
The \textsc{Observe} label exposes the earlier pre-event interval where a precursor cue is already legible, but the evidence is still insufficient for an immediate warning.
Ticks at or after $t^\star$ are discarded because they no longer represent a predictive alerting problem.
Negative clips are labelled \textsc{Silent} throughout.

The risk-cue onset $t^\circ$ is obtained with source-specific protocols: DoTA inherits its native anomaly interval, DADA-2000 uses manually reviewed risk-cue onset annotations, Nexar uses an optical-flow-based estimate, and DAD provides only clip-level labels with no usable $t^\circ$.
Sources without a usable $t^\circ$ contribute zero \textsc{Observe} ticks and split their usable pre-event interval between \textsc{Silent} and \textsc{Alert} only.
The full per-source protocol and the resulting class distribution are reported in Appendix~\ref{app:bench_split}.

\begin{table*}[t]
\centering
\footnotesize
\setlength{\tabcolsep}{6pt}
\renewcommand{\arraystretch}{0.92}
\begin{tabular*}{\textwidth}{@{\extracolsep{\fill}} l l rr rr @{}}
\toprule
\textbf{Source} & \textbf{Domain}
& \multicolumn{2}{c}{\textbf{Train}}
& \multicolumn{2}{c}{\textbf{Val / Held-out}} \\
\cmidrule(lr){3-4}
\cmidrule(lr){5-6}
& & Clips & Ticks & Clips & Ticks \\
\midrule
Nexar Collision & dashcam & 1{,}500 & 40{,}190 & 667 & 6{,}721 \\
DoTA            & dashcam & 2{,}949 & 29{,}763 & 326 & 3{,}256 \\
DAD             & dashcam & 1{,}157 & 4{,}628  & 127 & 508 \\
DADA-2000       & dashcam & 800     & 5{,}640  & 99  & 664 \\
\midrule
\textbf{In-domain total}
& --
& \textbf{6{,}406}
& \textbf{80{,}221}
& \textbf{1{,}219}
& \textbf{11{,}149} \\
\midrule
ADAS-TO-Critic & dashcam & -- & -- & 221 & -- \\
ACCIDENT       & synthetic CCTV & -- & -- & 2{,}211 & 17{,}224$^{\dagger}$ \\
\bottomrule
\end{tabular*}
\caption{\textbf{\name-Bench composition.}
In-domain sources are split by video and converted to 1\,Hz ticks with 8-frame observation windows.
ADAS-TO-Critic and ACCIDENT are held out for evaluation.
$^{\dagger}$Pre-accident ticks only.}
\label{tab:bench_sources}
\end{table*}


\subsection{Model Architecture}
\label{sec:method_architecture}
\name~contains a VLM belief extractor, a DangerHead, a PolicyHead, and an event-gated decoder.
We instantiate the extractor with Qwen3-VL-4B~\citep{bai2025qwen3}.
The visual encoder and original backbone weights are frozen.
The language layers are adapted with LoRA~\citep{hu2022lora}.
We add two belief delimiters and three action tokens: <|BELIEF|>, </|BELIEF|>, <SILENT>, <OBSERVE>, <|ALERT|>.
For each input, the VLM is trained to emit a short belief span followed by a frame-level action token: 
\[
\langle\textsc{|BELIEF|}\rangle \; w_1,\ldots,w_n \; \langle\textsc{/|BELIEF|}\rangle \; \langle\textsc{action}\rangle.
\]
The belief span contains concise safety evidence.
The special tokens make the belief region deterministic, so hidden states can be extracted without parsing free-form text.
Let $h_j^{(\ell)}$ be the hidden state at token position $j$ and layer $\ell$.
For frame $f$, let $o_f$ and $c_f$ be the positions of $\langle\textsc{B}\rangle$ and $\langle\textsc{/B}\rangle$.
We obtain a frame-level belief by span-pooling over a small set of upper layers $\mathcal{L}$:
\begin{equation}
\label{eq:belief_pooling}
z_t^{(f)}
=
\big\Vert_{\ell\in\mathcal{L}}
\left(
\frac{1}{c_f-o_f-1}
\sum_{j=o_f+1}^{c_f-1} h_j^{(\ell)}
\right).
\end{equation}
The sequence $Z_t=\{z_t^{(1)},\ldots,z_t^{(8)}\}$ forms the visual belief state.
We also read a decision register $r_t^{(f)}$ from the closing belief token.
This design is motivated by recent evidence that transformer hidden states can encode belief-like information in partially observed sequential processes~\citep{hu2025the}.
DangerHead estimates visual risk from $Z_t$.
It applies a shared frame-level MLP and aggregates the eight frame beliefs with learned-query attention.
It outputs per-frame danger scores $d_t^{(1:8)}$, a clip-level perception summary $S_t$, and a clip-level danger score $d_t^{\mathrm{clip}}$.
PolicyHead predicts the alert action from the decision-register sequence, the perception summary, the per-frame danger scores, and the previous-action embedding.
Formally, it computes
\begin{equation}
\label{eq:policy_input}
u_t=
\left[
\operatorname{Temp}(r_t^{(1:8)})
\Vert
S_t
\Vert
d_t^{(1:8)}
\Vert
e(a_{t-1})
\right].
\end{equation}
\begin{equation}
\label{eq:policy_softmax}
\pi_t=\operatorname{softmax}(g_\psi(u_t)).
\end{equation}
During policy training, DangerHead is frozen.
This keeps risk perception separate from alert decision-making.
At deployment, a finite-state decoder converts $\pi_t$ into a tick-level action.
A direct SILENT$\rightarrow$ALERT transition is allowed only when $p(ALERT)$ exceeds a high-confidence threshold.
A separate event-gating module then converts the dense tick stream into sparse driver-facing alerts.


\subsection{Training and Inference}
\label{sec:method_training}
\name is trained in three stages.
First, we perform belief-format supervised fine-tuning.
The VLM is trained with next-token cross-entropy over the assistant response so that it emits eight belief spans and eight frame-level action tokens.
Only LoRA parameters and the new special-token embeddings are updated.
This stage creates a reproducible hidden-state interface for downstream alerting. Second, we cache belief features from the fine-tuned VLM and train the downstream heads.

DangerHead is trained with frame-level and clip-level binary cross-entropy using the continuous danger target.
After DangerHead is trained, it is frozen.
PolicyHead is then trained with three-way cross-entropy over SILENT, OBSERVE, and ALERT.
A transition regularizer discourages low-confidence jumps from SILENT directly to ALERT.
Third, we perform closed-loop refinement.
The action-conditioned sampler is activated during training.
The previous action is first teacher-forced, then mixed with the model prediction, and finally taken from the model rollout.
This curriculum exposes PolicyHead to the same adaptive sampling distribution used at inference time.
At test time,~\name~repeats a closed loop.
It samples an 8-frame window from $W(V_{\leq t},a_{t-1})$.
It extracts belief states with the VLM.
It estimates danger with DangerHead.
It predicts $\pi_t$ with PolicyHead.
It decodes the tick-level action with the finite-state decoder.
It emits sparse driver-facing alerts through the event-gating module.
The decoded action is fed back as $a_t$ for the next tick.

\section{Experiments and Results}
\label{sec:experiments}

\subsection{Evaluation Protocol}
\label{sec:eval_protocol}

We evaluate all methods on the \name-Bench validation split, containing 11{,}149 ticks, 1{,}219 videos, and 794 positive videos.
Each method produces a scalar alert score per tick, and all threshold-dependent metrics are computed at one operating threshold $\tau$ per method.
We report tick-level ranking metrics, AUROC and AP\textsubscript{tick}; video-level coverage, Recall\textsubscript{v}; operating-point metrics, Precision\textsubscript{t}, F1\textsubscript{t}, and balanced accuracy; and lead-time metrics, mTTA@2s and mTTA@4s.
Because \name-Bench is dominated by SILENT ticks, we use balanced accuracy rather than raw accuracy:
\[
\mathrm{BalAcc}=\frac{\mathrm{TPR}+\mathrm{TNR}}{2}.
\]
This avoids rewarding constant-SILENT predictors and better reflects the deployed alerting trade-off.

\subsection{Driver-Aware Utility Score}
\label{sec:daus}

Standard accident-anticipation evaluation often reports mAP@TTA~\citep{BaoICCV2021DRIVE}, which measures threshold-free temporal ranking across Time-To-Accident buckets.
However, a deployed alerting system must operate at a fixed threshold, avoid nuisance alerts, cover dangerous videos, and warn early.
We therefore report \textbf{DAUS} (Driver-Aware Utility Score), a geometric utility over ranking, coverage, precision, and timing:
\begin{equation}
\label{eq:daus}
\begin{aligned}
\mathrm{DAUS} &= (M R_v P_t U_t)^{1/4}, \ \ \
U_t = \min\!\left(\frac{\mathrm{mTTA}}{L_{\mathrm{alert}}},1\right),
\end{aligned}
\end{equation}
where $M=\mathrm{mAP@TTA}$, $R_v=\mathrm{Recall}_v$, $P_t=\mathrm{Precision}_t$, and $U_t$ is normalized lead-time utility.
The multiplicative form penalizes failure in any component, making DAUS sensitive to both missed hazards and disruptive false alerts.

\subsection{Comparison with Baselines}
\label{sec:main_results}

We compare~\name~with five representative baselines: ResNet50-LSTM~\citep{7780459}, R3D-18~\citep{8578773},~\citep{li2021improved}, Open-BADAS~\citep{goldshmidt2025badas}, and Gemini-2.5-Flash~\citep{googledeepmind2025gemini25flashlite}. More details are in Appendix. \ref{app:baselines}.

Table~\ref{tab:main_results} reports the main comparison on \name-Bench validation.
\name{} achieves the best overall operating-point performance, ranking first in AUROC, Recall\textsubscript{v}, F1\textsubscript{t}, AP\textsubscript{tick}, Precision\textsubscript{t}, balanced accuracy, and DAUS.
Compared with the strongest baseline, Open-BADAS, \name{} improves AUROC from 0.610 to 0.689, AP\textsubscript{tick} from 0.176 to 0.195, and DAUS from 0.4752 to 0.4878.
This suggests that the proposed belief-state extraction and tri-action alert policy improve both tick-level discrimination and deployment-oriented decision quality beyond strong video representations.
Notably, MViT-V2-S obtains the highest mAP@TTA but the lowest DAUS, indicating that threshold-free temporal ranking alone can misrepresent driver-facing alert utility when video recall and tick-level precision at the deployment threshold are poor.

\begin{table*}[t]
\centering
\footnotesize
\setlength{\tabcolsep}{3pt}
\resizebox{\textwidth}{!}{
\begin{tabular}{l rrrrrr rr r r}
\toprule
Method
& AUROC$\uparrow$
& Recall\textsubscript{v}$\uparrow$
& F1\textsubscript{t}$\uparrow$
& AP\textsubscript{tick}$\uparrow$
& Prec\textsubscript{t}$\uparrow$
& BalAcc$\uparrow$
& mTTA@2s$\uparrow$
& mTTA@4s$\uparrow$
& mAP@TTA$\uparrow$
& DAUS$\uparrow$ \\
\midrule
\textbf{\name}
& \textbf{0.689}
& \textbf{0.884}
& \textbf{0.297}
& \textbf{0.195}
& \textbf{0.188}
& \textbf{0.648}
& \textbf{1.4}
& 3.0
& 0.503
& \textbf{0.4878} \\
Open-BADAS
& 0.610
& 0.882
& 0.276
& 0.176
& 0.184
& 0.581
& 1.2
& 2.3
& 0.512
& 0.4752 \\
R3D-18
& 0.601
& 0.800
& 0.242
& 0.161
& 0.159
& 0.573
& 1.3
& 2.9
& 0.506
& 0.4559 \\
ResNet50-LSTM
& 0.581
& 0.800
& 0.218
& 0.147
& 0.138
& 0.541
& \textbf{1.4}
& 3.0
& 0.506
& 0.4439 \\
Gemini-2.5-Flash-Lite
& 0.568
& 0.712
& 0.220
& 0.153
& 0.173
& 0.554
& 1.1
& 2.1
& 0.504
& 0.4344 \\
MViT-V2-S
& 0.566
& 0.750
& 0.206
& 0.148
& 0.129
& 0.523
& \textbf{1.4}
& \textbf{3.1}
& \textbf{0.530}
& 0.4329 \\
\bottomrule
\end{tabular}
}
\caption{\textbf{Main results on \name-Bench validation.}
All threshold-dependent metrics in a row use the same operating threshold $\tau$.
}
\label{tab:main_results}
\end{table*}

\begin{figure}[!ht]
    \centering
    \includegraphics[width=0.85\linewidth]{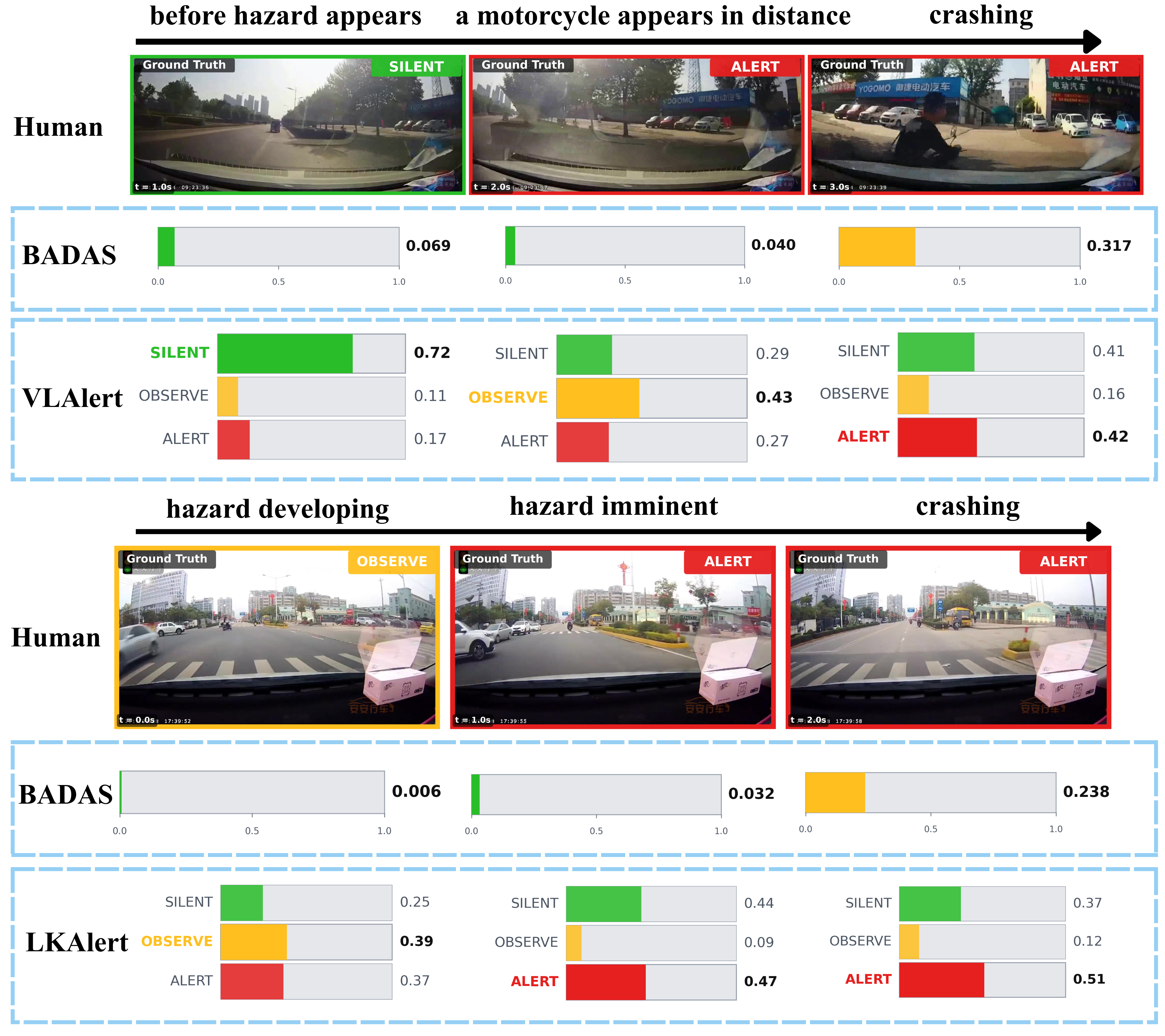}
    \caption{Performance comparison of~\name~and Open-BADAS. }
    \label{fig:results}
\end{figure}



\subsection{Ablation Studies}
\label{sec:ablation}

We ablate the two components that distinguish~\name~from a conventional binary VLM-based alerter: the structured \texttt{BELIEF} interface and the explicit OBSERVE action.

\begin{figure}[!h]
    \centering
    \includegraphics[width=1\linewidth]{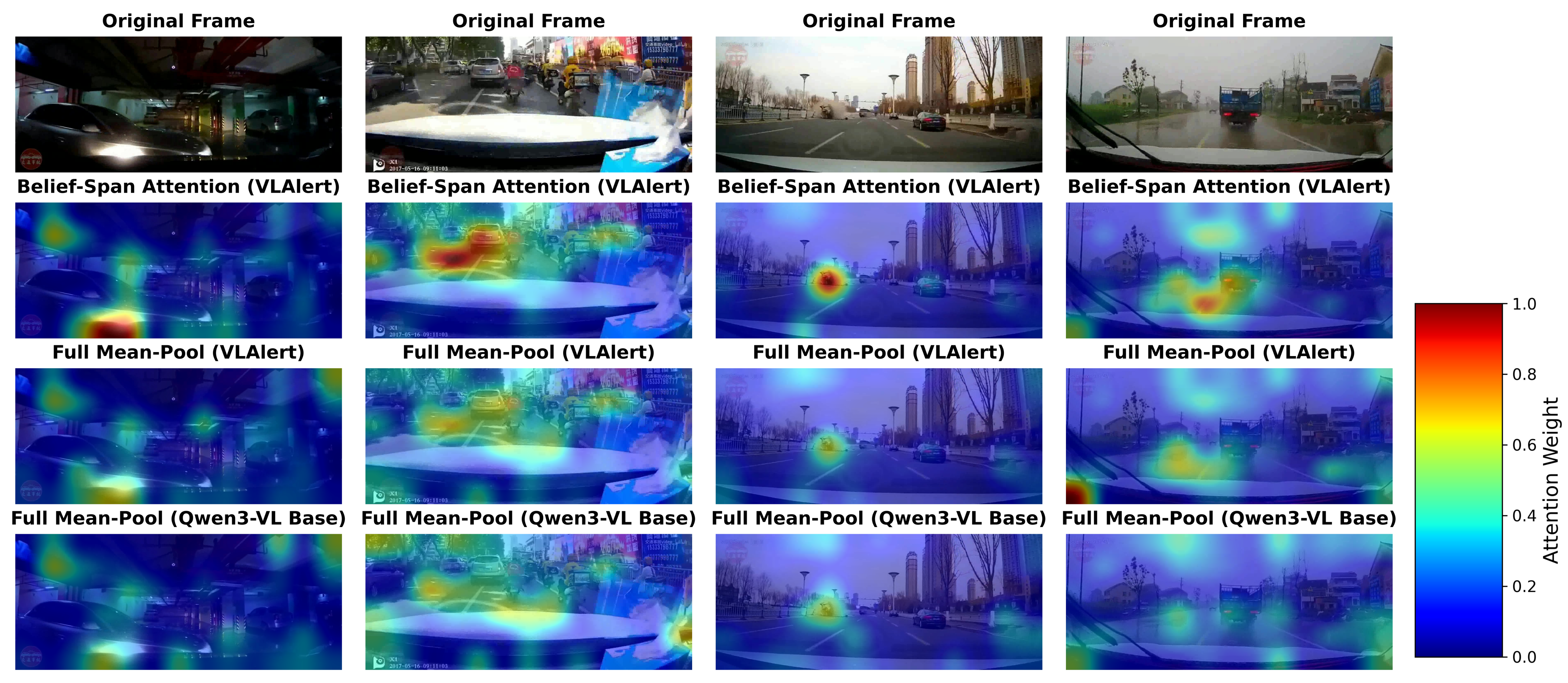}
    \caption{Comparing attention of the adaptive BELIEF-span and the default VLM output using mean-pool: Results show that BELIEF-span sees the whole scene with right focus on driving risk visual cues.}
    \label{fig:attention}
\end{figure}

\paragraph{BELIEF pooling.}

We freeze the SFT-tuned Qwen3-VL-4B backbone and compare three ways of extracting hidden-state features from the same generated response.
A 5-fold linear probe is trained for binary ALERT prediction on the \name-Bench validation split.
As shown in Table~\ref{tab:abl_belief}(a), pooling over the \texttt{BELIEF} span gives the strongest linear separability, with AP 0.459 and AUROC 0.894.
Using only the opening \texttt{BELIEF} token remains competitive, while pooling a same-length random response span drops AP to 0.204.
This supports the hypothesis that the structured \texttt{BELIEF} format localizes safety-relevant evidence in predictable hidden-state positions, rather than leaving it diffusely distributed across the generated text.

\begin{table*}[!h]
\centering
\footnotesize
\setlength{\tabcolsep}{7pt}
\renewcommand{\arraystretch}{1.08}
\resizebox{\textwidth}{!}{
\begin{tabular}{l rrr @{\hspace{1.5em}} l rrr}
\toprule
\multicolumn{4}{c}{\textbf{(a) BELIEF pooling} (linear probe, 5-fold CV)}
&
\multicolumn{4}{c}{\textbf{(b) OBSERVE action space} (3 seeds, F1$^\star$-optimal $\tau$)} \\
\cmidrule(lr){1-4}
\cmidrule(lr){5-8}
Pooling strategy
& AP $\uparrow$
& AUROC $\uparrow$
& F1 $\uparrow$
&
Action space
& AP$_t$ $\uparrow$
& AUROC$_t$ $\uparrow$
& F1$_t$ $\uparrow$ \\
\midrule
\textbf{\texttt{BELIEF} range (ours)}
& \textbf{0.459}
& \textbf{0.894}
& \textbf{0.442}
&
\textbf{3-class S/O/A (ours)}
& \textbf{0.184}
& \textbf{0.734}
& \textbf{0.256} \\

\texttt{BELIEF} open-token only
& 0.419
& 0.882
& 0.414
&
Binary, OBSERVE$\to$SILENT
& 0.175
& 0.705
& 0.255 \\

random-span control
& 0.204
& 0.760
& 0.222
&
Binary, OBSERVE$\to$ALERT
& 0.179
& 0.726
& 0.254 \\
\bottomrule
\end{tabular}
}
\caption{\textbf{Ablations on \name-Bench validation.}
(a) \texttt{BELIEF}-aligned pooling yields substantially stronger alert separability than random-span pooling, indicating that the structured span localizes safety-relevant evidence in predictable hidden-state positions.
(b) Keeping OBSERVE as a separate action improves tick-level ranking over collapsing it into either SILENT or ALERT, suggesting that OBSERVE provides useful supervision for uncertain pre-alert states.}
\label{tab:abl_belief}
\end{table*}

\paragraph{OBSERVE supervision.}
We retrain PolicyHead with three label spaces while keeping the backbone and DangerHead features fixed.
Table~\ref{tab:abl_belief}(b) shows that the three-action formulation obtains the best AP$_t$, AUROC$_t$, and F1$_t$ across three seeds.
It improves AP$_t$ by 5.4\% and 3.1\% over binary classifier, indicating that OBSERVE provides a useful supervision signal for ranking uncertain pre-alert states rather than acting as a redundant intermediate label.

\smallskip

The paper also validates the \name{} performance against two unseen datasets, ADAS-TO, which tests if the model aligns with human takeover actions before crashes happen, and another synthesized dataset in CARLA from a different roadside view (Appendix \ref{app:validation}).

\section{Conclusion}
\label{sec:conclusion}

We presented \name{}, a vision–language framework that reframes dashcam driver alerting as a tri-action sequential policy over SILENT, OBSERVE, and ALERT, with OBSERVE serving as an evidence-gathering deferment rather than an intermediate confidence bin. The framework consists of three components: a unified per-tick benchmark, VLAlert-Bench, paired with the DAUS metric that jointly scores ranking, video coverage, precision, and lead-time utility; a tri-action policy that adapts observation frequency; and a VLM belief-extraction interface that emits focused safety spans to downstream heads. On VLAlert-Bench, VLAlert obtains the best overall operating-point utility across all baselines. Our ablations further show that pooling hidden states over the BELIEF span yields substantially stronger alert separability than pooling random spans of equal length, and that retaining OBSERVE as a separate action improves tick-level ranking over collapsing it into either SILENT or ALERT.


\bibliography{references}  

\appendix

\section{Validation in Unseen Datasets}
\label{app:validation}

\subsection{ADAS-TO-Critic: Generalization to Human Takeovers}
\label{app:adasto}

\begin{table*}[!h]
\centering
\footnotesize
\setlength{\tabcolsep}{7pt}
\renewcommand{\arraystretch}{0.88}
\begin{tabular}{lcccccc}
\toprule
Method
& F1$\uparrow$
& R@10s$\uparrow$
& R@5s$\uparrow$
& Lead@10s$\uparrow$
& Lead@5s$\uparrow$
& DAUS$\uparrow$ \\
\midrule
\textbf{\name{} (Ours)}
& \textbf{0.686}
& \textbf{0.941}
& \textbf{0.887}
& 6.12
& \textbf{3.88}
& \textbf{0.520} \\
Open-BADAS
& 0.585 & 0.810 & 0.742 & 5.83 & 3.52 & 0.515 \\
R3D-18
& 0.418 & 0.602 & 0.498 & \textbf{7.02} & 3.75 & 0.485 \\
MViT-V2-S
& 0.355 & 0.588 & 0.498 & 6.65 & 3.74 & 0.495 \\
ResNet50-LSTM
& 0.385 & 0.579 & 0.498 & 6.66 & 3.74 & 0.479 \\
Gemini-2.5-Flash-Lite (zs)
& 0.013 & 0.036 & 0.032 & 5.12 & 3.43 & 0.393 \\
\bottomrule
\end{tabular}
\caption{\textbf{ADAS-TO-Critic evaluation.}
The split contains 221 expert-reviewed takeover clips centered at $t=10$\,s.
R@10s/R@5s measure pre-takeover alert coverage; Lead@10s/Lead@5s report the mean first-alert lead time.
\textit{zs} denotes zero-shot prompting.}
\label{tab:adasto-test}
\end{table*}

We further evaluate \name{} on \textbf{ADAS-TO-Critic}, a held-out set of 221 expert-reviewed dashcam clips from production ADAS logs. Each clip is centered at a confirmed human takeover at $t=10$\,s. This setting tests whether an alert policy trained on accident and anomaly data can transfer to naturalistic takeover events, where the driver's intervention provides an external behavioral signal of perceived risk.

Table~\ref{tab:adasto-test} reports the results. \name{} achieves the best F1, R@10s, R@5s, Lead@5s, and DAUS. In particular, it detects 88.7\% of takeover clips within the last 5\,s before intervention, compared with 74.2\% for Open-BADAS. The improvement in DAUS indicates that \name{} not only fires more often before takeovers, but does so with better timing under the deployment utility criterion.

\begin{figure}[!h]
\centering
\includegraphics[width=0.85\linewidth]{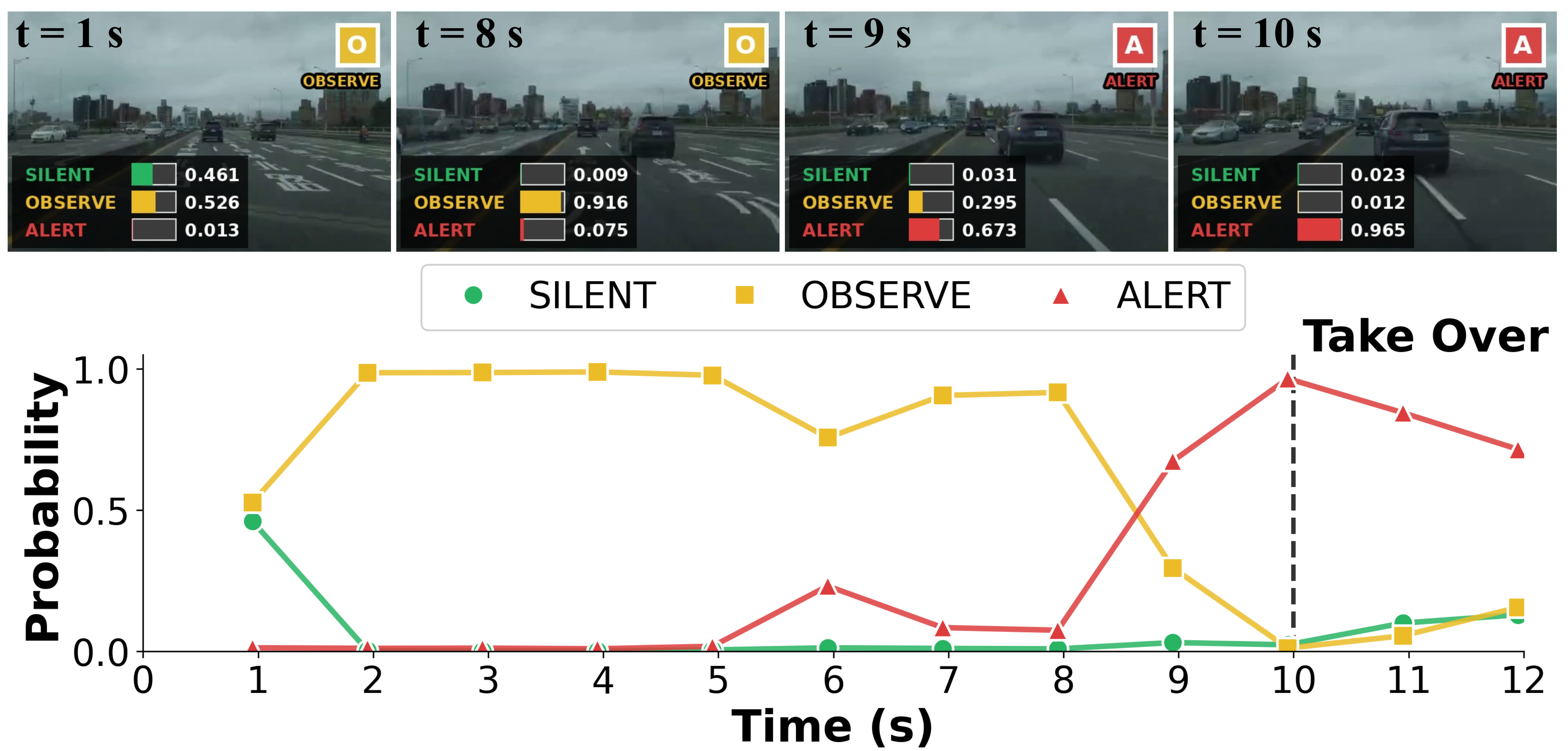}
\caption{\textbf{Example prediction on ADAS-TO-Critic.}
Top: selected frames with predicted action and class probabilities. Bottom: 1\,Hz probability trace over time; the dashed line marks the human takeover at $t=10$\,s. }
\label{fig:adasto-example}
\end{figure}

Figure~\ref{fig:adasto-example} shows a representative case. \name{} remains in OBSERVE while risk is developing, then switches to ALERT one second before the recorded takeover and peaks at takeover time. This illustrates the intended OBSERVE$\rightarrow$ALERT behavior: defer under uncertainty, then warn sharply when intervention becomes imminent.

\subsection{Out-of-Distribution Generalization on Roadside Accidents}
\label{sec:carla_accident}

We next evaluate whether \name{} transfers beyond ego-view dashcam data.
We use a roadside CARLA accident benchmark with 2{,}211 collision clips captured from fixed roadside cameras.
This setting changes both viewpoint and scene statistics: the camera is static, the view is third-person, and the accident often unfolds across an intersection rather than from the ego vehicle's perspective.
Since all clips are positive, AP and AUROC are not meaningful; we instead report per-clip alert rate and mean time-to-accident (mTTA) over the full pre-accident interval, the last 5\,s, and the last 2\,s before impact.

\begin{figure}[!h]
\centering
\includegraphics[width=0.99\linewidth]{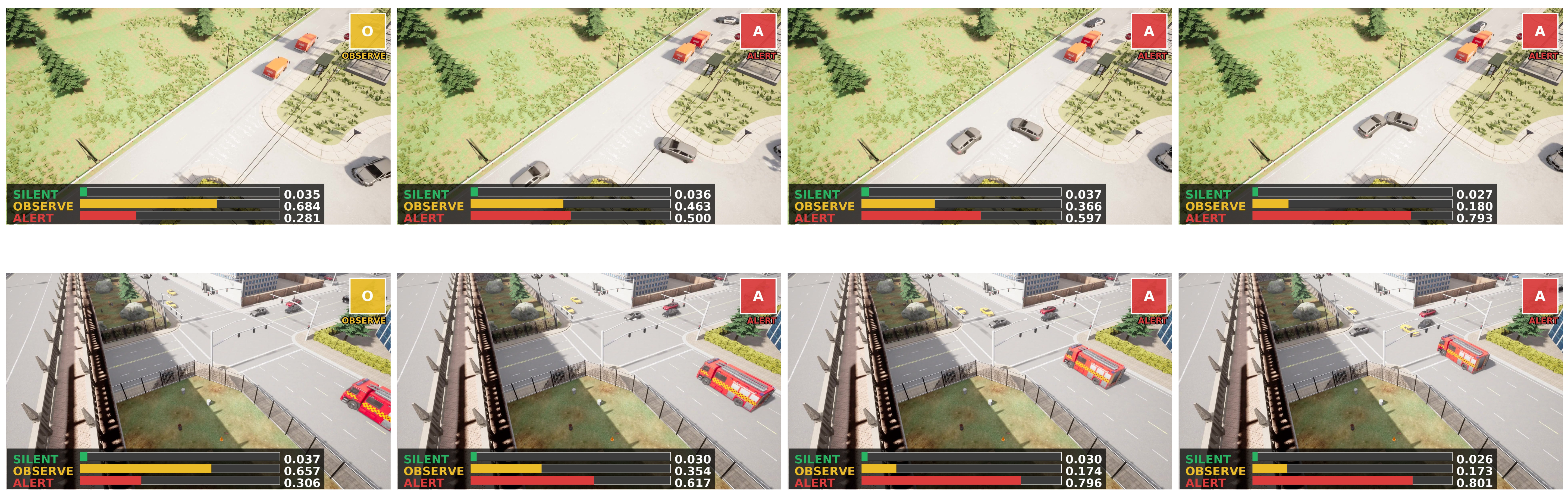}
\caption{Roadside CARLA accident example.
A fixed roadside camera captures an unprotected left-turn collision.
\name{} escalates from OBSERVE to ALERT before impact and sustains the alert, while Gemini remains in SILENT.}
\label{fig:accident_example}
\end{figure}

Table~\ref{tab:carla_accident} shows that \name{} transfers well to this unseen viewpoint.
It alerts on 88.9\% of clips over the full pre-accident window with a mean lead time of 7.31\,s, and still fires on 74.8\% of clips within the last 2\,s before impact.
In contrast, zero-shot Gemini-2.5-Flash-Lite rarely issues an alert under the same three-action protocol.
Figure~\ref{fig:accident_example} gives a representative example: \name{} escalates from OBSERVE to ALERT several seconds before impact, while Gemini remains silent.

\begin{table*}[t]
\centering
\footnotesize
\setlength{\tabcolsep}{6pt}
\renewcommand{\arraystretch}{0.86}
\begin{tabular*}{\textwidth}{@{\extracolsep{\fill}} l cc cc cc c @{}}
\toprule
Method
& \multicolumn{2}{c}{Full}
& \multicolumn{2}{c}{Last 5\,s}
& \multicolumn{2}{c}{Last 2\,s}
& DAUS$_{+}$$\uparrow$ \\
\cmidrule(lr){2-3}
\cmidrule(lr){4-5}
\cmidrule(lr){6-7}
& Rate$\uparrow$ & mTTA$\uparrow$
& Rate$\uparrow$ & mTTA$\uparrow$
& Rate$\uparrow$ & mTTA$\uparrow$
& \\
\midrule
\textbf{\name{} (Ours)}
& \textbf{88.9\%} & \textbf{7.31}
& \textbf{83.1\%} & \textbf{4.18}
& \textbf{74.8\%} & 1.49
& \textbf{0.945} \\
Gemini-2.5-Flash-Lite
& 2.1\% & 5.59
& 1.5\% & 3.38
& 0.4\% & \textbf{1.50}
& 0.510 \\
\bottomrule
\end{tabular*}
\caption{\textbf{Roadside CARLA accident evaluation.}
Rate is the percentage of clips with at least one ALERT; mTTA is the mean first-alert lead time in seconds. $\mathrm{DAUS}_{+}$
  is the degenerate utility defined in
  Appendix~\ref{app:carla_accident}.}
\label{tab:carla_accident}
\end{table*}

\section{Benchmark Construction Details}
\label{app:bench_split}

\paragraph{Event-time extraction.}
Each source is converted to a common tick-level stream using its native annotation format.
For Nexar Collision, the event time $t^\star$ is read from the released metadata.
For DoTA, we use the annotated anomaly interval $[t_{\mathrm{start}}, t_{\mathrm{end}})$.
For DADA-2000, $t^\star$ is the accident time from the clip annotation.
DAD does not provide reliable per-frame event timestamps, so its clip-level labels are propagated to all usable ticks.
ADAS-TO-Critic and ACCIDENT are held out for evaluation.

\paragraph{Tick generation.}
For each clip, we sample a 1\,Hz tick stream.
A tick is anchored at time $t_f$ and exposes the eight frames immediately preceding that anchor.
The tick label is the action label assigned to its anchor frame under Eq.~\ref{eq:label_rule}.
For collision-style corpora, ticks after the alert window are discarded because they no longer test predictive alerting.
Negative clips are labelled SILENT throughout.

\begin{figure}[!h]
\centering
\includegraphics[width=\linewidth]{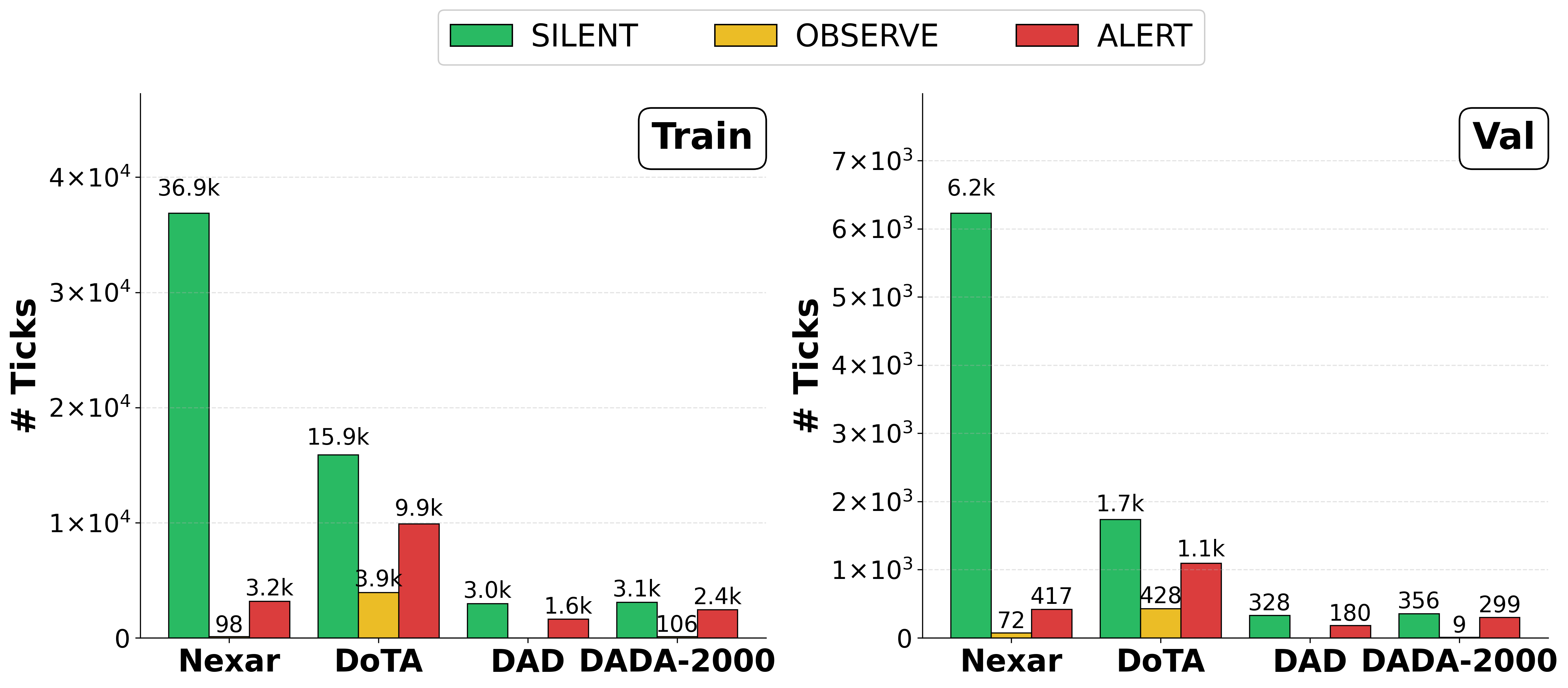}
\caption{\textbf{\name-Bench class distribution.}
Per-source tick counts for train and validation after applying the label rule and post-event filtering.
Bars show absolute tick counts for SILENT, OBSERVE, and ALERT.}
\label{fig:bench_distribution}
\end{figure}

\paragraph{Label distribution.}
Figure~\ref{fig:bench_distribution} reports the train and validation labels distributions after filtering.
Nexar contributes most SILENT ticks, while DoTA contributes most ALERT ticks and supplies nearly all OBSERVE ticks.
DAD is the smallest source, and DADA-2000 provides a denser mix of normal and alert ticks.

\paragraph{Risk-cue onset and OBSERVE labels.}
The base labelling rule in Eq.~\ref{eq:label_rule} assigns SILENT and ALERT labels from the event time.
The OBSERVE label is added as an intermediate pre-event state.
For a clip with event time $t^\star$, we define $t^\circ\leq t^\star$ as the first time at which the visual precursor of the event becomes legible.
Ticks in $[t^\circ,t^\star)$ are relabelled as OBSERVE, replacing the SILENT label that would otherwise be assigned before the event.
Because the four in-domain sources provide different levels of temporal annotation, $t^\circ$ is obtained with source-specific rules.

\begin{itemize}[leftmargin=*,topsep=2pt,itemsep=1pt]
    \item \textbf{DoTA.}
    DoTA provides frame-level anomaly annotations and anomaly categories.
    We use the first annotated anomaly frame as the risk-cue onset $t^\circ$, and use the contact or critical-proximity frame as $t^\star$ when available.
    This rule is applied automatically to all DoTA clips.
    As a result, DoTA supplies most of the OBSERVE ticks in \name-Bench.

    \item \textbf{DADA-2000.}
    DADA-2000 provides accident-cause text and driver-attention maps.
    Two authors manually annotated $t^\circ$ for all DADA-2000 clips used in our training and validation splits.
    We define $t^\circ$ as the first frame where the at-risk actor becomes visible and matches the salient entity described by the accident-cause annotation.
    Annotators used a frame-level review interface with attention overlays.
    Agreement on a 100-clip audit subset was Cohen's $\kappa=0.79$.
    Disagreements were resolved by joint review.

    \item \textbf{Nexar Collision.}
    Nexar provides collision time but no anomaly interval or attention map.
    We therefore estimate $t^\circ$ automatically from pre-event motion.
    For each clip, we compute dense optical-flow magnitude over a 1\,s rolling window and mark $t^\circ$ as the last pre-event frame where the maximum flow magnitude exceeds the clip-level 90th percentile.
    Clips without a clear pre-event flow peak receive no OBSERVE label and remain SILENT until the event.

    \item \textbf{DAD.}
    DAD provides clip-level binary labels but no reliable frame-level event time.
    We therefore do not assign risk-cue onset times for DAD.
    DAD contributes only SILENT and ALERT ticks.
\end{itemize}

\paragraph{Annotation quality control.}
The manual annotation steps were conducted by two authors with experience in traffic-safety data annotation.
A third author reviewed a random 10\% sample from each batch.
The residual disagreement rates were 4.1\% for DoTA cleanup and 6.3\% for DADA-2000 risk-cue onset marking.
All disagreements were resolved by joint adjudication.
We did not use crowd workers, since identifying the first safety-relevant visual precursor requires domain knowledge and consistent temporal judgement.

The resulting OBSERVE distribution matches the annotation strength of each source.
In the training split, the 4{,}152 OBSERVE ticks consist of 3{,}948 from DoTA, 106 from DADA-2000, 98 from Nexar, and 0 from DAD, as reflected in Figure~\ref{fig:bench_distribution}.

\section{\name{} Training Details}
\label{app:vlalert_training}

This appendix describes the data pipeline, training stages, and inference protocol of \name{}.
All stages are run with bf16 mixed precision on a single NVIDIA RTX~5090 GPU with 32\,GB memory.
The full pipeline takes approximately 80 GPU-hours.

\subsection{Training-Data Construction}
\label{app:vlalert_data}

\paragraph{Sample unit.}
Each training sample follows the tick format of \name-Bench: an 8-frame observation window anchored at time $t_f$, paired with an action label
$y\in\{SILENT,OBSERVE,ALERT\}$.
The tick label is derived from the event time $t^\star$ using the labelling rule in Eq.~\ref{eq:label_rule}.
For belief-format supervised fine-tuning, we also assign per-frame labels $y_1,\ldots,y_8$ by applying the same rule to each frame timestamp.

\paragraph{Belief-span text.}
Each frame is paired with a short safety-evidence span enclosed by the \texttt{<|BELIEF|>} and \texttt{</|BELIEF|>} delimiters.
When a dataset provides native textual annotations, such as accident causes or attention regions, we reuse them as safety-evidence text.
For datasets without rich text annotations, we use the base instruction-tuned Qwen3-VL-4B model to generate concise safety captions with the prompt:
\begin{quote}
\small
Describe the safety-relevant evidence in this dashcam frame in no more than 20 words. Mention only hazards, agents, and traffic state.
\end{quote}
Generated captions are filtered for length, generic content, and obvious failures.
Failed captions are replaced with short category-based templates derived from the action label.
For OBSERVE samples, risk-cue onset annotations are used to bias the belief text toward the developing hazard rather than generic scene description.

\paragraph{Assistant response format.}
For every 8-frame tick, the supervised target is an eight-line response:
\begin{small}
\begin{verbatim}
<|BELIEF|> belief_1 </|BELIEF|> <action_1>
<|BELIEF|> belief_2 </|BELIEF|> <action_2>
...
<|BELIEF|> belief_8 </|BELIEF|> <action_8>
\end{verbatim}
\end{small}
where \texttt{<action\_i>} is one of \texttt{<SILENT>}, \texttt{<OBSERVE>}, or \texttt{<|ALERT|>}.
This fixed format makes the belief boundaries deterministic and supports the span-pooling operation in Eq.~\ref{eq:belief_pooling}.

\paragraph{Prompt template.}
The prompt used for belief-format training and feature extraction is:
\begin{small}
\begin{verbatim}
[SYSTEM]
You are a driving-safety analyst. You will be shown eight frames
from a driving video. For each frame, identify the safety-relevant
evidence and decide the appropriate alert level for the driver.

[USER]
Frames: <img_1><img_2>...<img_8>
For each of the eight frames, output one line in the form
<|BELIEF|> evidence </|BELIEF|> <action>
where <action> is one of <SILENT>, <OBSERVE>, or <|ALERT|>.
\end{verbatim}
\end{small}
We avoid specifying a fixed intra-window sampling rate in the prompt because the closed-loop sampler changes the temporal spacing of the eight frames according to the previous action.

\paragraph{Final training set.}
Applying this pipeline to the in-domain split of \name-Bench yields 80{,}221 training ticks and 11{,}149 validation ticks.
Each tick contains eight frames and the corresponding belief/action supervision.

\subsection{Stage 1: Belief-Format Supervised Fine-Tuning}
\label{app:vlalert_sft}

\paragraph{Trainable parameters.}
We start from Qwen3-VL-4B-Instruct.
The visual encoder and the original language-model backbone are frozen.
LoRA adapters are attached to the linear projections in the language transformer blocks, including query, key, value, output, up, gate, and down projections.
We add five new tokens to the tokenizer:
\texttt{<|BELIEF|>}, \texttt{</|BELIEF|>}, \texttt{<SILENT>}, \texttt{<OBSERVE>}, and \texttt{<|ALERT|>}.
Only the LoRA weights and the embeddings of the new tokens are trained.
LoRA uses rank 64, scaling factor $\alpha=128$, and dropout 0.05.

\paragraph{Training-data composition.}
Stage~1 uses the 80{,}221 training ticks from the in-domain \name-Bench split.
Each tick contains eight frames and therefore contributes eight supervised
\texttt{<|BELIEF|>}\,...\,\texttt{</|BELIEF|>} spans, yielding 641{,}768 belief spans in total.
Table~\ref{tab:vlalert_belief_sources} summarizes the supervision sources.
When native textual annotations are available, such as DADA-2000 accident-cause and attention text or DoTA anomaly categories, we reuse them as safety-evidence descriptions.
For the remaining frames, we generate belief text offline with GPT-5.4 using the prompt:
\begin{quote}
\small
Describe the safety-relevant evidence in this frame in no more than 20 words. Mention only hazards, agents, and traffic state.
\end{quote}
The generated text is filtered for length, generic captions, and obvious failures.
Residual failures are replaced with short class-templated fallbacks derived from the per-frame label $y_i$.
No additional free-form manual captioning is used.

\begin{table*}[t]
\centering
\footnotesize
\setlength{\tabcolsep}{5pt}
\renewcommand{\arraystretch}{0.95}
\begin{tabular*}{\textwidth}{
@{\extracolsep{\fill}}
l
r
r
r
>{\raggedright\arraybackslash}p{0.22\textwidth}
>{\raggedright\arraybackslash}p{0.25\textwidth}
@{}}
\toprule
\textbf{Source}
& \textbf{Clips}
& \textbf{Ticks}
& \textbf{Belief spans}
& \textbf{Belief origin}
& \textbf{Example snippet} \\
\midrule
Nexar Collision
& 1{,}500
& 40{,}190
& 321{,}520
& GPT-5.4 caption
& ``\textit{lead truck brakes hard, gap closing fast}'' \\

DoTA
& 2{,}949
& 29{,}763
& 238{,}104
& GPT-5.4 caption + anomaly label
& ``\textit{pedestrian crossing from right curb}'' \\

DAD
& 1{,}157
& 4{,}628
& 37{,}024
& GPT-5.4 caption
& ``\textit{motorcycle weaving between lanes ahead}'' \\

DADA-2000
& 800
& 5{,}640
& 45{,}120
& Native cause + attention text
& ``\textit{rear-end risk; lead car decelerating sharply}'' \\
\midrule
\textbf{Total}
& \textbf{6{,}406}
& \textbf{80{,}221}
& \textbf{641{,}768}
& --
& -- \\
\bottomrule
\end{tabular*}
\caption{\textbf{Belief supervision for Stage~1 SFT.}
Each training tick provides eight per-frame belief spans.
Native textual annotations are reused when available; otherwise, concise safety-evidence captions are generated offline with GPT-5.4 and filtered before training.}
\label{tab:vlalert_belief_sources}
\end{table*}

\paragraph{Objective.}
The model is trained with next-token cross-entropy over the assistant response only.
System, user, and image tokens are masked out.
Belief-text tokens use weight 1.0, while delimiter and action tokens use weight 2.0 to reduce collapse toward the majority SILENT label.
Formally, for the target token sequence $z_{1:T}$ and token weights $w_t$, the Stage~1 loss is
\[
\mathcal{L}_{\mathrm{SFT}}
=
-\frac{1}{\sum_t w_t}
\sum_{t=1}^{T}
w_t \log p_\theta(z_t \mid z_{<t}, X),
\]
where $w_t=0$ for masked prompt and image tokens, $w_t=1$ for belief-text tokens, and $w_t=2$ for delimiter and action tokens.

\paragraph{Optimization.}
We use AdamW with learning rate $5\times 10^{-5}$, weight decay 0.05, cosine decay, 3\% warm-up, gradient clipping at 1.0, and bf16 precision.
The micro-batch size is one tick, with gradient accumulation over 16 steps, giving an effective batch size of 16 ticks.
We train for three epochs with maximum sequence length 4096 and gradient checkpointing enabled.
Stage~1 takes approximately 60 GPU-hours.

\subsection{Stage 2: Belief Caching and Downstream Heads}
\label{app:vlalert_heads}

\paragraph{Belief cache.}
After Stage 1, the LoRA-adapted VLM is frozen.
We run it over all in-domain ticks and extract hidden states from each generated \texttt{<|BELIEF|>} span.
For each frame $f\in\{1,\ldots,8\}$, we span-pool the hidden states between \texttt{<|BELIEF|>} and \texttt{</|BELIEF|>} and concatenate the last four transformer layers, following Eq.~\ref{eq:belief_pooling}.
This yields a 10{,}240-dimensional belief vector per frame.
We also cache a decision register from the closing \texttt{</|BELIEF|>} token.
Feature caching takes approximately 6 GPU-hours.

\paragraph{DangerHead.}
DangerHead estimates visual risk from the eight frame-level belief vectors.
It first applies a shared two-layer MLP,
$10240\rightarrow1024\rightarrow512$, with GELU, LayerNorm, and dropout 0.1.
A learned-query attention pool aggregates the eight frame features into a clip-level summary, and linear heads output both per-frame danger scores and a clip-level danger score.
The loss is the sum of per-frame and clip-level binary cross-entropy losses using the ALERT indicator as the target.
We train with AdamW, learning rate $10^{-3}$, weight decay $10^{-4}$, batch size 256 ticks, and early stopping on validation AP.
Training takes approximately 3 GPU-hours.
DangerHead is frozen after this stage.

\paragraph{PolicyHead.}
PolicyHead predicts the three-way action distribution.
Its input consists of the decision-register sequence, the clip-level perception summary, the per-frame danger scores, and a 16-dimensional embedding of the previous action.
The temporal module is a two-layer Transformer with 4 heads and hidden dimension 256, followed by a three-way linear classifier.
The objective is class-balanced cross-entropy over
$\{SILENT,OBSERVE,ALERT\}$, plus a transition regularizer that discourages low-confidence direct jumps from SILENT to ALERT.
We train with AdamW, learning rate $5\times10^{-4}$, batch size 128 ticks, and 40 epochs.
Training takes approximately 3 GPU-hours.

\subsection{Stage 3: Closed-Loop Refinement}
\label{app:vlalert_closedloop}

The final stage exposes PolicyHead to the same action-conditioned observation distribution used at inference time.
The sampler $W(V_{\leq t},a_{t-1})$ chooses the next 8-frame window according to the previous action:
a wide sparse window after SILENT, a medium dual-resolution window after OBSERVE, and a dense short-range window after ALERT.
The previous action follows a three-phase curriculum:
\begin{enumerate}[leftmargin=*,topsep=2pt,itemsep=1pt]
    \item teacher forcing in the first epoch, using the ground-truth previous action;
    \item scheduled sampling in the second epoch, linearly mixing ground-truth and model-predicted previous actions;
    \item full rollout in the final epoch, using the model's own previous prediction.
\end{enumerate}
The VLM backbone and DangerHead are frozen; only PolicyHead is updated.
We use learning rate $10^{-4}$ and the same transition regularizer as in Stage 2.
This stage takes approximately 8 GPU-hours.

\subsection{Finite-State Decoder and Event Gating}
\label{app:vlalert_fsm}

The finite-state decoder and event-gating module are rule-based and contain no learned parameters.
Their hyperparameters are calibrated once on the \name-Bench validation split and then frozen for all evaluations.

\paragraph{Finite-state decoder.}
The decoder maintains the previous decoded state
$s_{t-1}\in\{SILENT,OBSERVE,ALERT\}$,
which is also passed to the action-conditioned sampler.
At tick $t$, it receives the PolicyHead distribution
$\pi_t=(p_{\textsc{S}},p_{\textsc{O}},p_{\textsc{A}})$
and outputs the current action $a_t$.
If $s_{t-1}=SILENT$, a direct transition to
ALERT is allowed only when
$p_{\textsc{A}}>\tau_{\mathrm{jump}}$.
Otherwise, the decoder switches to OBSERVE when
$p_{\textsc{O}}+p_{\textsc{A}}>0.5$ and remains in
SILENT otherwise.
If $s_{t-1}\in\{OBSERVE,ALERT\}$, the decoder simply takes
$\arg\max_a \pi_t(a)$.
The only threshold in this decoder is $\tau_{\mathrm{jump}}$.
We sweep
$\tau_{\mathrm{jump}}\in\{0.30,0.35,\ldots,0.95\}$
on the validation split and select the value that maximizes DAUS under the same operating constraint used in the main experiments.
The selected value is $\tau_{\mathrm{jump}}=0.60$, which is used unchanged on all held-out datasets.

\paragraph{Event gating.}
The decoded tick-level actions are converted into sparse driver-facing alerts by a deterministic event gate.
An alert event is emitted only after at least $H=2$ consecutive
ALERT ticks, which suppresses isolated one-tick flickers.
After an alert is emitted, the system enters a refractory window of
$R=3$ seconds during which no new alert event can be emitted.
The values $H=2$ and $R=3$ are selected by the same validation sweep used for $\tau_{\mathrm{jump}}$.

\paragraph{Deployment use.}
The decoder and event gate add no trainable parameters and no training cost.
The calibrated triple
$(\tau_{\mathrm{jump}},H,R)=(0.60,2,3)$
is fixed for all reported results, including ADAS-TO-Critic and ACCIDENT, with no dataset-specific re-tuning.

\subsection{Inference Protocol}
\label{app:vlalert_inference}

At inference time, \name{} runs at 1\,Hz.
Given the previous action $a_{t-1}$, the action-conditioned sampler selects an 8-frame window $X_t$.
The VLM produces belief spans and action tokens, and the hidden states inside the belief spans are pooled into frame-level belief vectors.
DangerHead estimates per-frame and clip-level risk.
PolicyHead then predicts $\pi_t$ over SILENT, OBSERVE, and ALERT.
A finite-state decoder blocks low-confidence SILENT$\rightarrow$ALERT transitions unless $p(ALERT)>0.6$.
Last, an event-gating module merges consecutive ALERT ticks into sparse driver-facing alerts.
The measured latency is approximately 180\,ms per tick on the RTX~5090, within the 1\,Hz update budget.


\section{Baseline Architectures and Training Details}
\label{app:baselines}

All learned baselines are trained on the same \name-Bench in-domain training split of 80{,}221 ticks from 6{,}406 clips and validated on the same 11{,}149-tick validation split.
Training uses bf16 mixed precision on a single NVIDIA RTX~5090.
For each learned baseline, we tune one operating threshold on the validation set using the same calibration protocol as \name{}.
Zero-shot baselines are not trained.

\subsection{ResNet50-LSTM}

\paragraph{Architecture.}
A ResNet-50 image encoder is applied independently to each of the eight frames.
The 2048-dimensional global-average-pooled feature from each frame is projected to 512 dimensions and passed to a single-layer LSTM with hidden size 512.
The final hidden state is fed to a two-layer MLP,
$512\rightarrow256\rightarrow2$, producing binary alert logits.

\paragraph{Training.}
Frames are resized to $224\times224$ and normalized with ImageNet statistics.
The ResNet-50 backbone is frozen during a two-epoch warm-up and then unfrozen.
We optimize binary cross-entropy with AdamW.
The learning rate is $10^{-4}$ for the LSTM and classifier and $10^{-5}$ for the ResNet backbone.
Weight decay is $10^{-4}$.
We use cosine decay, 5\% warm-up, batch size 16 ticks, and class-balanced sampling.
Training runs for 12 epochs, and the best checkpoint is selected by validation AP.
Training takes approximately 18 GPU-hours.

\subsection{R3D-18}

\paragraph{Architecture.}
We use a Kinetics-pretrained R3D-18 backbone.
The eight frames in a tick are stacked into a short video clip and resized/cropped to $112\times112$.
The global average-pooled 512-dimensional feature is passed to a linear binary classifier.

\paragraph{Training.}
The full backbone is trained from the first epoch.
We use SGD with momentum 0.9, learning rate $10^{-3}$, weight decay $5\times10^{-4}$, cosine decay, 5\% warm-up, and batch size 32 clips.
Data augmentation includes horizontal flips, brightness jitter, and small temporal jitter.
Training runs for 20 epochs, with the best checkpoint selected by validation AP.
Training takes approximately 14 GPU-hours.

\subsection{MViT-V2-S}

\paragraph{Architecture.}
We use a Kinetics-pretrained MViT-V2-S backbone.
Each 8-frame tick is resized and center-cropped to $224\times224$.
The pooled CLS feature is passed through LayerNorm and a linear binary classifier.

\paragraph{Training.}
The full backbone is fine-tuned from epoch 0.
We use AdamW with learning rate $10^{-4}$, weight decay $5\times10^{-2}$, layer-wise learning-rate decay 0.75, cosine decay, 10\% warm-up, drop-path rate 0.1, batch size 32 clips, and class-balanced sampling.
Training runs for 25 epochs, and the best checkpoint is selected by validation AP.
Training takes approximately 22 GPU-hours.

\subsection{Open-BADAS}

\paragraph{Architecture.}
Open-BADAS uses a frozen V-JEPA2 ViT-L/16 video encoder.
Each 8-frame tick is resampled to 16 frames at $256\times256$.
The encoder output is passed to a lightweight collision head consisting of attention pooling followed by a two-layer MLP for binary collision logits.

\paragraph{Training.}
The V-JEPA2 backbone is frozen, and only the collision head is optimized.
We use AdamW with learning rate $5\times10^{-4}$, weight decay $10^{-2}$, cosine decay, 5\% warm-up, batch size 16 ticks, and class-balanced sampling.
Training runs for 30 epochs, with the best checkpoint selected by validation AP.
Training takes approximately 6 GPU-hours.

\subsection{Gemini-2.5-Flash-Lite}

\paragraph{Setup.}
Gemini-2.5-Flash-Lite is used as a zero-shot multimodal baseline.
For each tick, the eight frames are encoded as JPEG images and sent in one multimodal request.
We use temperature 0 and a 100-token output cap.
No fine-tuning or in-context examples are used.

\paragraph{Prompt.}
The prompt is:
\begin{small}
\begin{verbatim}
You are a driving-safety classifier. You will see eight frames
from a vehicle-mounted dashcam. Decide the appropriate alert level
for the last frame:
  SILENT  - normal driving, no hazard developing
  OBSERVE - risk developing, but more evidence is needed
  ALERT   - a collision or near-collision is imminent or in progress

Respond only with a JSON object:
{"action":"SILENT|OBSERVE|ALERT",
 "p_silent":0..1,
 "p_observe":0..1,
 "p_alert":0..1}
where the three probabilities sum to 1.
\end{verbatim}
\end{small}

For the roadside-CCTV evaluation, the first sentence is changed to:
\begin{quote}
\small
You will see eight frames from a fixed roadside camera overlooking a road or intersection.
\end{quote}
All other prompt content is unchanged.
The predicted class is the argmax over the three probabilities, and the scalar alert score is $p_{ALERT}$.

\paragraph{Score-to-action conversion.}
All non-\name{} learned baselines, except Gemini-2.5-Flash-Lite, output a scalar danger score $s\in[0,1]$ at each tick rather than an explicit alert action.
To evaluate them under the same threshold-dependent protocol as \name{}, we convert each score into a binary decision in $\{SILENT,ALERT\}$ using a single global threshold.
For each baseline, we sweep $\tau\in\{0.01,0.02,\ldots,0.99\}$ on the \name-Bench validation split and select the threshold $\tau^\star$ that maximizes balanced accuracy under the same video-recall constraint used for \name{}.
The selected threshold is then frozen for all subsequent evaluations:
a tick is labelled ALERT if $s\geq\tau^\star$ and SILENT otherwise.
For held-out datasets such as ADAS-TO-Critic, we reuse the same $\tau^\star$ without re-tuning, so the reported results reflect the fixed operating point that would be used in deployment.


\section{ADAS-TO-Critic Dataset Details}
\label{app:adasto_critic}

\paragraph{Source and curation.}
ADAS-TO-Critic is a held-out, test-only set built from the public ADAS-TO collection.
ADAS-TO provides 285 candidate critical dashcam clips collected during real production ADAS operation.
Each clip follows a fixed temporal structure: the vehicle is under L2 ADAS control for the first 10\,s, the human driver takes over at $t=10$\,s, and the vehicle is manually driven for the remaining 10\,s.
Two authors independently reviewed all 285 clips and removed cases where the takeover was not primarily safety-motivated, such as comfort-driven overrides, construction-zone protocol takeovers, or disengagements caused by known system limits rather than external hazards.
The remaining 221 clips form ADAS-TO-Critic.
Inter-reviewer agreement on the keep/discard decision is Cohen's $\kappa=0.84$; disagreements were adjudicated by a third author.
All retained clips preserve the original takeover anchor at $t^\star=10$\,s.

\paragraph{Why this split tests alerting.}
ADAS-TO-Critic provides a behavioral reference for driver alerting.
The takeover at $t=10$\,s is an external signal that the driver perceived the situation as requiring immediate intervention.
This makes the split different from accident-only evaluation: the target is not collision occurrence itself, but whether the alert model fires before a human intervention that was motivated by perceived safety risk.
The split is test-only and shares no clips with the in-domain training corpora described in Appendix~\ref{app:bench_split}.

\paragraph{Tick generation.}
Each 20\,s clip is converted into a 1\,Hz tick stream.
We keep ticks anchored at $t_f\in\{1,2,\ldots,12\}$\,s, where each tick exposes the eight frames immediately preceding $t_f$.
Ticks after $t_f>12$\,s are discarded because they fall inside the post-takeover manual-driving portion and no longer test predictive alerting.
Evaluation is binary at the clip level: a method either fires at least one ALERT before takeover or it does not.
No OBSERVE sub-label is used for scoring on this split.

\paragraph{Evaluation metrics.}
For each clip, we record the first ALERT prediction before takeover, denoted $\tau_{\mathrm{fire}}$ when it exists.
We report:
\begin{itemize}[leftmargin=*,topsep=2pt,itemsep=1pt]
    \item \textbf{R@10s}: the fraction of clips with at least one ALERT in $[0,10]$\,s;
    \item \textbf{R@5s}: the fraction of clips with at least one ALERT in $[5,10]$\,s;
    \item \textbf{Lead@10s}: the mean lead time $10-\tau_{\mathrm{fire}}$ over clips that fire in $[0,10]$\,s;
    \item \textbf{Lead@5s}: the same lead time, restricted to clips that fire in $[5,10]$\,s.
\end{itemize}
F1 is computed from the binary fire/no-fire decision over the full pre-takeover window.
No threshold is tuned on ADAS-TO-Critic.
All learned baselines use the operating threshold selected on \name-Bench validation, and \name{} uses the fixed decoder and event-gating parameters from Appendix~\ref{app:vlalert_fsm}.


\section{CARLA Accident OOD Evaluation Details}
\label{app:carla_accident}

\begin{figure}[!h]
\centering
\includegraphics[width=0.97\linewidth]{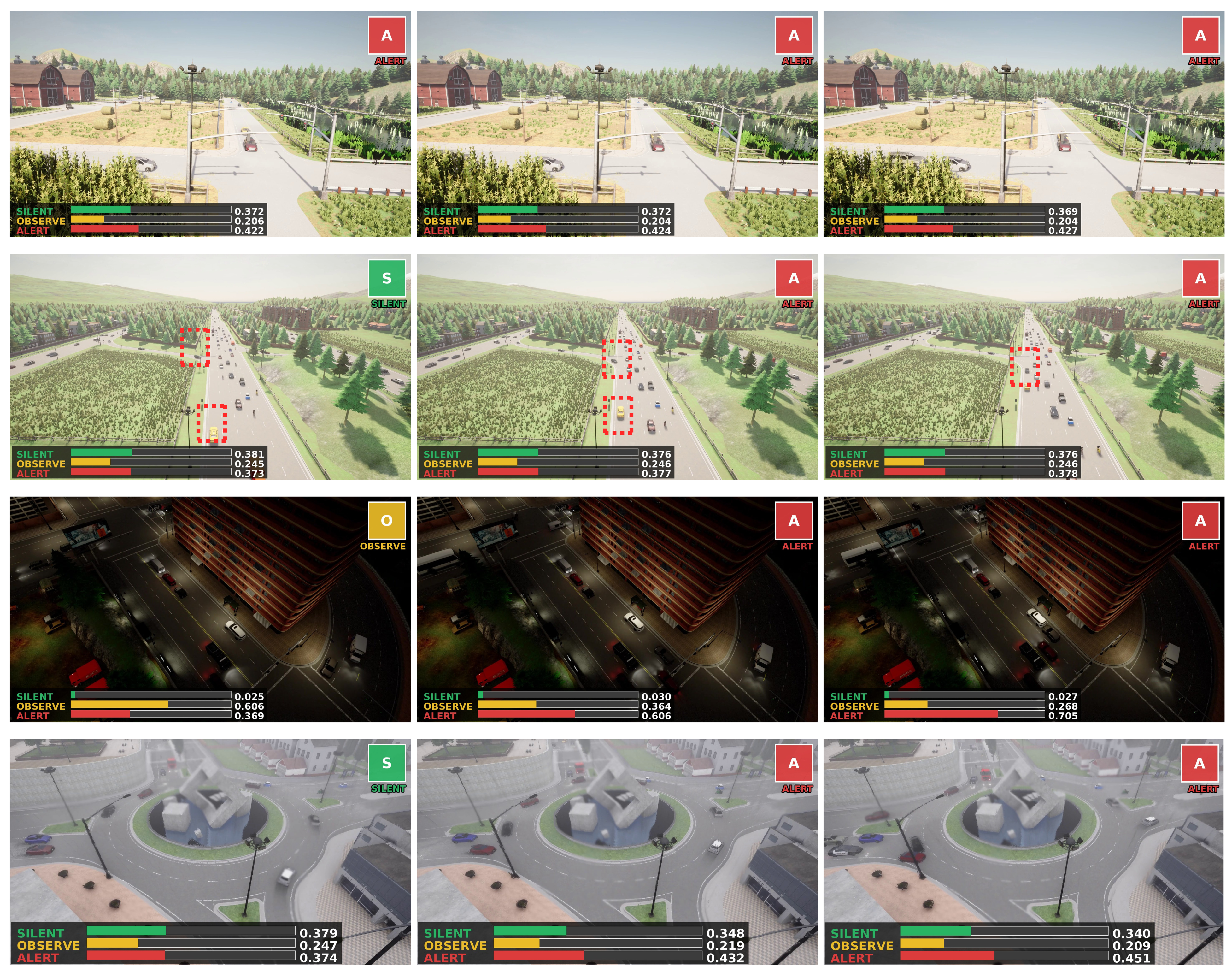}
\caption{VLAlert performance on roadside Carla accident dataset.}
\label{fig:appendix_accident}
\end{figure}

\paragraph{Benchmark composition.}
The CARLA accident split is used only for out-of-distribution evaluation.
It contains 2{,}211 clips, each ending in a labelled collision with known accident time $t^\star$.
All clips are captured by a static roadside camera at 20\,fps.
The accident labels cover five categories: rear-end, head-on, sideswipe, t-bone, and single-vehicle.
Compared with the in-domain dashcam data, this split changes the viewpoint, field of view, camera motion, and accident geometry.
Figure~\ref{fig:appendix_accident} shows the class distribution and an example failure case.

\paragraph{Evaluation protocol.}
All clips in this split are positive collision clips, so standard binary
AP, AUROC, and the multiplicative DAUS in Eq.~\ref{eq:daus} are not
well-defined for this evaluation.  In particular, there are no negative
clips to rank against or to estimate a false-alert burden, and
clip-level precision becomes degenerate once a method fires.  We
therefore report positive-set alerting metrics:
\begin{itemize}[leftmargin=*,topsep=2pt,itemsep=1pt]
    \item \emph{Alert rate}: the fraction of clips with at least one
    ALERT inside the evaluation window;
    \item \emph{mTTA}: the mean time-to-accident of the first
    ALERT, computed over clips that fire;
    \item \emph{DAUS$_+$}: a positive-only utility summarising coverage
    and lead time,
    \[
    \mathrm{DAUS}_{+}
    =
    \frac{1}{2}U_{+}
    +
    \frac{1}{2}(1-U_{-}),
    \qquad
    U_{+}
    =
    \mathrm{Rate}_{\mathrm{full}}
    \cdot
    \min\!\left(
    \frac{\mathrm{mTTA}_{\mathrm{full}}}{5},1
    \right).
    \]
\end{itemize}
Here $U_{+}$ is the product of full-window alert coverage and capped
lead-time utility.  Since this split contains no negative clips, we set
$U_{-}=0$.  Thus, DAUS$_+$ preserves the same $[0,1]$ range as
Eq.~\ref{eq:daus} and is monotone in both alert coverage and lead time.
However, it is not directly comparable to the validation DAUS in
Section~\ref{sec:main_results}; it should be read only as a compact
OOD summary of the per-window alert-rate and mTTA results.

We report three windows: the full pre-accident interval, the last
5\,s before impact, and the last 2\,s before impact.  All decisions use
argmax over the three-action output together with the finite-state
decoder and event-gating parameters fixed on \name-Bench validation
(Appendix~\ref{app:vlalert_fsm}).  No threshold is tuned on this split.

\paragraph{Baseline.}
We compare with Gemini-2.5-Flash-Lite under a zero-shot CCTV-aware prompt.
Each tick is represented by eight consecutive frames, and the model is asked to choose among SILENT, OBSERVE, and ALERT.
Gemini was evaluated on 937 clips due to scoring coverage; \name{} was evaluated on all 2{,}211 clips.

\paragraph{Per-type results.}
Table~\ref{tab:carla_accident_full} reports results by accident type.
\name{} performs best on sideswipe, rear-end, and head-on crashes, where the risk is visible through sustained closing or lateral motion.
T-bone cases are harder because the crossing vehicle often enters the camera view shortly before impact.
Single-vehicle crashes are the main failure mode: without a second actor, the useful signal is mainly vehicle kinematics, which is less represented in the current belief supervision.

\begin{table*}[!h]
\centering
\footnotesize
\setlength{\tabcolsep}{5pt}
\renewcommand{\arraystretch}{0.92}
\begin{tabular*}{\textwidth}{@{\extracolsep{\fill}} l l r c c c c @{}}
\toprule
\textbf{Method} 
& \textbf{Type} 
& \textbf{Clips}
& \textbf{Full/mTTA} 
& \textbf{Last 5\,s/mTTA} 
& \textbf{Last 2\,s/mTTA} 
& \textbf{DAUS} \\
\midrule
\multirow{6}{*}{\textbf{\name{} (Ours)}}
& rear-end       & 794 & \textbf{94.0 / 7.10}  & \textbf{87.0 / 4.25}  & \textbf{78.2 / 1.49} & \textbf{0.970} \\
& head-on        & 588 & \textbf{93.5 / 6.72}  & \textbf{84.7 / 3.82}  & \textbf{77.6 / 1.47} & \textbf{0.968} \\
& sideswipe      & 405 & \textbf{100.0 / 9.85} & \textbf{99.8 / 4.50}  & \textbf{98.8 / 1.52} & \textbf{1.000} \\
& t-bone         & 358 & \textbf{69.8 / 5.17}  & \textbf{65.1 / 4.21}  & \textbf{47.8 / 1.44} & \textbf{0.849} \\
& single-vehicle & 66  & \textbf{21.2 / 6.28}  & \textbf{16.7 / 3.92}  & \textbf{7.6 / 1.34}  & \textbf{0.606} \\
\cmidrule(l){2-7}
& \textbf{Overall} 
& \textbf{2{,}211}
& \textbf{88.9 / 7.31}
& \textbf{83.1 / 4.18}
& \textbf{74.8 / 1.49}
& \textbf{0.945} \\
\midrule
\multirow{6}{*}{Gemini-2.5-Flash-Lite}
& rear-end       & 794 & 1.9 / 5.75 & 1.0 / 2.22 & 0.6 / 1.75 & 0.510 \\
& head-on        & 588 & 1.6 / 7.41 & 0.8 / 4.28 & 0.0 / --   & 0.508 \\
& sideswipe      & 405 & 0.9 / 5.45 & 0.9 / 2.45 & 0.9 / 1.45 & 0.505 \\
& t-bone         & 358 & 5.3 / 3.91 & 5.3 / 3.63 & 0.8 / 1.05 & 0.521 \\
& single-vehicle & 66  & 0.0 / --   & 0.0 / --   & 0.0 / --   & 0.500 \\
\cmidrule(l){2-7}
& Overall 
& 937
& 2.1 / 5.59
& 1.5 / 3.38
& 0.4 / 1.50
& 0.510 \\
\bottomrule
\end{tabular*}
\caption{\textbf{CARLA accident OOD evaluation by accident type.}
Each window reports \textbf{Rate / mTTA}, where Rate is per-clip alert coverage in percent and mTTA is the mean first-alert lead time in seconds.
Dashes indicate that no ALERT was fired in that window.}
\label{tab:carla_accident_full}
\end{table*}

\paragraph{Limitations.}
The main failure modes are single-vehicle loss-of-control and late-entry t-bone cases.
The former lacks a second interacting actor, while the latter may provide less than two seconds of visible pre-impact evidence.
Both failures suggest that future versions should include longer temporal context, explicit ego-motion or vehicle-kinematic belief fields, and a small amount of CCTV-style adaptation data.

\end{document}